# An Automatic Ontology Generation Framework with An Organizational Perspective

Samaa Elnagar
Virginia Commonwealth University
elnagarsa@vcu.edu

Victoria Yoon
Virginia Commonwealth University
vyyoon@vcu.edu

Manoj A. Thomas
University of Sydney
manoj.thomas@sydney.edu.au

## Abstract

*Ontologies have been known for their powerful semantic representation of knowledge. However, ontologies cannot automatically evolve to reflect updates that occur in respective domains. To address this limitation, researchers have called for automatic ontology generation from unstructured text corpus. Unfortunately, systems that aim to generate ontologies from unstructured text corpus are domain-specific and require manual intervention. In addition, they suffer from uncertainty in creating concept linkages and difficulty in finding axioms for the same concept. Knowledge Graphs (KGs) has emerged as a powerful model for the dynamic representation of knowledge. However, KGs have many quality limitations and need extensive refinement. This research aims to develop a novel domain-independent automatic ontology generation framework that converts unstructured text corpus into domain consistent ontological form. The framework generates KGs from unstructured text corpus as well as refine and correct them to be consistent with domain ontologies. The power of the proposed automatically generated ontology is that it integrates the dynamic features of KGs and the quality features of ontologies.*

## 1. Introduction

Ontologies have been used as a model for knowledge storage and representation [1]. Characteristics of good ontology are: *memory, dynamism, polysemy, and automation* [2]. In an ideal scenario, systems must be capable of generating and enriching ontologies automatically. However, most ontologies are generated manually by ontology engineers who are familiar with the theory and practice of ontology construction [3]. The goal of automatic ontology generation is to convert new knowledge into ontological form by enabling related processing techniques, such as semantic search and retrieval [4]. Automatic ontology generation will significantly reduce the labor cost and time required to build ontologies [5].

Most of the current automatic ontology generation systems convert existing structured knowledge (e.g., database schemas and XML documents) into ontological formats [6]. However, approaches to convert unstructured text corpus into ontological format have not been fully developed. Moreover, existing approaches are domain-specific and require manual intervention to create domain rules and patterns [2]. Similar to ontologies, Knowledge Graphs (KGs) encode structured information of entities and their relations into a graphical form or a directed graph $G = (C, R)$, where $C$ is the set of vertices and $R$ is the set of edges that symbolizes a relationship between two concepts in a graph [7]. However, there is a significant difference between ontologies and KGs that are important to note. First, from a practical viewpoint, KGs are powerful in many aspects; however, their quality and reliability are questionable. Second, there is usually a trade-off between coverage and correctness of KGs, which could be perilous for certain business problems. Third, from a theoretical perspective, the trustworthiness of KGs has not been established [8], particularly in cases of organizations that delegate high priority to data quality and systems reliability.

This research aims to develop a domain-independent automatic ontology generation framework that enable organizations to generate ontological form from unstructured text corpus. The study is fueled by the lack of fully-automated domain-independent ontology generation systems that address common data quality issues. The framework utilizes refined KGs to be mapped and tailored to fit into target domain ontologies. The generated ontologies benefit from KGs' features and avoid quality issues traditionally associated with automatic ontology generation. In addition to enabling organizations to store and retrieve new knowledge in ontological RDF format, the study also shows how the framework can facilitate *interoperability* to efficiently employ knowledge across multiple domains. It is to be noted that generated ontologies are in the basic triple RDF format and their hierarchical structure (i.e., the OWL format) is beyond the scope of the paper. However, generating ontologies from refined KGs will not only overcome the limitations of ontologies such as *data integration* and *evolution* but also take advantage of the benefits of KGs such as *timeliness*.

The contribution of this paper can be summarized as:
1. The design of an automatic ontology generation framework from unstructured knowledge sources that can be used across various domains.
2. KGs alignment with reference ontologies after refining them in terms of *correctness, completeness, and consistency* with target domain ontologies.
3. The development of criteria for KG correction and consistency check.

## 2. Literature Review

Due to enhancements in machine learning and Natural Language Processing (NLP) algorithms, many studies have addressed the generation of ontology from unstructured knowledge. Syntactic pattern methodologies and rule-based approaches have been used extensively in ontology engineering [9]. However, those approaches require manually crafted sets of rules or patterns to represent knowledge, making them narrow in scope and domain dependent. Authors in [10] and [11] presented ontology generation systems from plain text using predefined dictionary, statistical, and NLP techniques. The two approaches target the medical domain specifically. Additionally, their approaches require extensive labor costs to construct patterns and maintain the comprehensive dictionaries.

The system in [12] used Wikipedia texts to extract concepts and relations for ontology construction. They used a supervised machine learning technique which required huge effort for manual labeling and validation for data. An Alzheimer ontology generation system was built in [13]. The system used controlled vocabulary along with linked data to build the ontology based on Text2Onto system by combining machine learning approaches with part-of-speech (POS) tagging. Unfortunately, involvement of domain experts is needed during the development process.

Alobaidi et al. asserted [3] the need for automatic and domain-independent ontology generation methods. They identified biomedical concepts using Linked life Data (LOD) and linked medical knowledge-bases and applied semantic enrichment to enrich concepts. Breadth-First Search (BFS) algorithm was used to direct the LOD repository to create precise well-defined ontology. However, this approach targets the medical domain and the framework is trained only with linked biomedical ontologies. Further, the quality of the generated ontologies is neither evaluated nor checked for error and consistency with domain ontologies. The system in [14] automatically constructed an ontology from a set of text documents using WordNet, but no details were provided on how the terms are extracted and no qualitative assessment is provided.

Kong et. al. [15] designed a domain-specific automatic ontology system based on WordNet. The approach is highly dependent on the quality of the starting knowledge resource. User intervention is also necessary to avoid incompatible concepts. In [16], dictionary parsing mechanisms and discovery methods were used for acquiring domain-specific concepts. The developed framework is considered a semi-automatic ontology acquisition system for mining ontologies from textual resources. The framework also depends on technical dictionaries for building a concept taxonomy for the target domain.

Meijer et al. [17] developed a framework for the generation of a domain taxonomy from a text corpora. The framework employed a disambiguation step for both the extracted taxonomy and the reference ontology used for evaluation. In addition, the subsumption method was used for hierarchy creation. However, the scope of this study was only to build a taxonomy of concepts and relations with minimal focus on relations between instances. Further, the system has very low semantic precision and recall because of improper relation representation.

To summarize, we conclude that most of the approaches used for automatic ontology generation from unstructured text corpus are domain-specific, demonstrating the need for domain independent ontology-generation methods [3, 4]. Additionally, few systems used unstructured text from external heterogeneous sources [17, 18]. Human intervention was also required in one or more tasks. Further, few approaches take into consideration the quality of the generated ontology. Moreover, issues such as timeliness, evolution, and integration have not been discussed. To fill this gap in the literature, our study aims to develop a fully automatic, domain independent ontology generation framework using various types of unstructured text corpus. Quality issues are the core consideration in the design of our system. In our method, we will focus on the relationships between different instances with reference to the structure of reference ontologies.

## 3. What are Knowledge Graphs?

A knowledge graph is used mainly to describe real world entities and their interrelations organized in a graph [19]. It is considered a dynamically growing semantic network of facts about things. Some argue that KGs are somehow superior to ontologies and provide additional features such as timeliness and scalability [20]. A knowledge graph $G$ consists of schema graph $Gs$, data graph $Gd$ and the relations R between $Gs$ and $Gd$, denoted as $G = <Gs, Gd, R>$. KGs schema do not necessarily contain all concepts and relations as the domain ontology. In contrast, KGs are generated based on the concepts found in the source corpus. For example, Figure 1.a. shows a KG for IMDB reviews. In the figure, movies are connected to actors, directors and genres. We can easily imply that the IMDB reviewer Rita (the beige colored node in the middle) likes Charles Chaplin as an actor and a director. However, the ontological representation for such graph would be the expansion of the composing ontologies shown as in Figure 1.b. Expanding the basic ontologies will include all concepts and attributes in each ontology will result in unnecessary concepts representation.

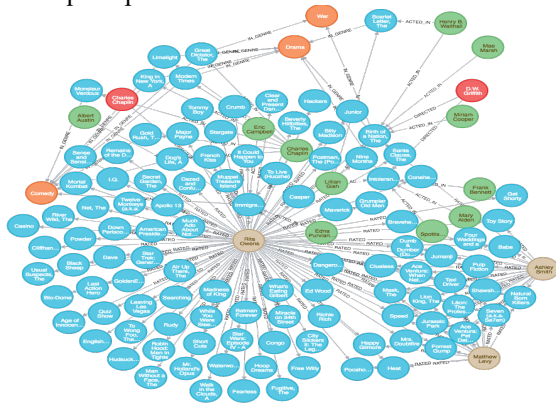

**Figure 1.a**[1]: Knowledge Graph for IMDB reviews and the basic equivalent ontologies

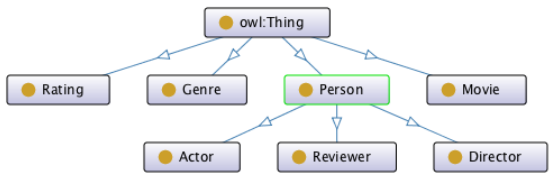

**Figure 1.b** The main equivalent ontologies representation for retail store web application

## 4. A Comparison between KGs and Ontologies

A KG is considered a dynamic or problem specific ontology [21]. KGs are domain-independent methods for knowledge representation, while ontologies are known for representing domain knowledge [22]. So, in KGs the number of instances statements is far larger than that of schema level statements. The focus of knowledge graphs is the instance (A-box) level more than the concept (T-box) level. While ontologies focus on building schematic taxonomies of concepts and relations for a certain domain [23], KG schema is rather shallow, at a small degree of formalization without hierarchical structure. Most KGs follow the *Open World Assumption (OWA)* which states that KGs contain only true facts and the non-observed facts can be either false or just missing. On the other hand, most of ontologies follow a domain-specific approach or the *Closed World Assumption (CWA)* that assumes that the facts not contained in the domain are false [24].

*Scalability* is "the ability of a system to be enlarged to accommodate growth" [25]. KGs are very scalable as shown in Figure 1.a. *Reliability* as a concept depends on the availability of sources [26] and therefore ontologies availability of data is higher than KGs. The automatic *construction* and *maintenance* of KGs faces substantial challenges. Maintenance of KGs depends on manual user feedback which is burdensome, subjective and difficult [27].

*Timeliness* measures how up-to-date data is relative to a specific task [28]. Since most updates in ontologies are done manually by domain experts [29], There is a time that the ontology will be incomplete. In contrast, KGs are generated at runtime and the data is current. *Evolution* is very likely in KGs for several reasons: (i) KGs represent dynamic resources, and (ii) the entire graph can change or disappear [30]. On the other hand, ontologies need domain expert's intervention to evolve and it is usually a daunting and costly process.

*Licensing* is defined as "the granting of permission for a consumer to re-use a dataset under defined conditions" [31]. *Licensing* is a new quality dimension not considered for relational databases [32]. KGs should contain a license or clear legal terms so that the content can be (re)used. *Interoperability* is "the usage of relevant vocabularies for a particular domain" [33] such that different systems can exchange information. *Interoperability* is a main issue in KGs and some argue that KGs may create inconsistencies with many information systems [27, 34].

*Relevancy* refers to "the provision of information which is in accordance with the task at hand" [35]. Because KGs are multi domain graphs, knowledge about certain domain might be superficial. Unlike ontologies, they contain detailed descriptions of concepts and relations for a specific domain. Data integration in ontologies is challenging specially when it comes to extending the knowledge beyond the domain knowledge [36, 37]. *Data integration* in case of KG might lead to duplication of instances and referential conflicts. Therefore, KG refinement is crucial. KGs are essential for real time processing such real-time recommendations and fraud detection [38].

---

[1] The figure is generated using NEO4j sandbox

The size of KGs usually is far larger than the size of ontologies [39]. Although extensively in use, KGs are hard to compare against each other in a given setting [40]. Unlike KGs, there are many tools that are used to compare different ontologies [41]. *Computational performance* concerns become more important as KGs become larger. Typical performance measures are runtime measurements, as well as memory consumption [42]. Since knowledge graphs explicitly identify all concepts and their relationships to each other, they are inherently explainable which is not the case with ontologies [43].

KGs have a higher degree of *agility,* the rate of knowledge change*,* than ontologies because of their dynamicity and continuous evolution [29]. *Redundancy* refers to the duplication of relations, attributes or instances [27]. KGs might be generated on the fly, so they are prune to duplication of instances. KGs are connecting instances visually, so they are human friendly. KGs have changed the nature of many ML techniques such as the graph-convolutional neural network [44]. A comparison between ontologies and KGs is illustrated in Table 1.

**Table 1:** A Comparison between KGs and ontologies

| Criteria | KG | Ontology | Source |
| --- | --- | --- | --- |
| Assumption | *OWA* | *CWA* | [24] |
| Size | Massive | Relatively small | [39] |
| Scalability | Very scalable | Limited scalability | [25] |
| Scope | Problem specific | Domain specific | [21] |
| Real-time | Generated at runtime | Limited real-time capability | [38] |
| Timeliness | Fresh | Outdated | [29] |
| Generation | Automatic | Mostly by humans | [3, 21] |
| Trustworthiness | Not very trustworthy | Trustworthy | [45] |
| Knowledge base type | More A-Box than T-Box | Usually more T-Box than A-Box | [23] |
| Markup language | RDF | RDF, OWL, OIL | [46] |
| Data Integration | Easily integrated | Hard to Integrate | [47] |
| Quality (Correctness, Completeness) | Questionable | High Quality | [27] |
| Agility | Dynamic | Static | [19] |
| Redundancy | Very likely | Not Likely | [29] |
| Reliability | Questionable | Reliable | [26] |
| Maintenance | Challenging | Burdensome | [48] |
| Evolution | Easy | Difficult | [30] |
| Security (licensing) | Questionable | Reasonable | [31] |
| Interoperability | Low | Moderate | [33] |
| Relevancy | Low | High | [35] |
| Computational Performance | Heavy | Light | [42] |
| Comparability | Very Hard | Achievable | [40] |
| Friendliness | Machine / human friendly | Machine / not human friendly | [43] |

## 4.1. Domain and Quality Constraints

Building knowledge graphs from scratch is a tedious proposition that require machine learning models to be trained with huge number of datasets in addition to strong NLP techniques and reasoning capabilities. However, third-parties solutions, such as IBM *Watson*, and *Neo4j* [49], offer knowledge graphs generation as on-demand services. However, some organizations may not trust third-party generated KGs owing to concerns of security, reliability and relevancy. Therefore, organizations should weigh the time and effort required to produce a knowledge graph against the value it receives from using third-parties KGs.

## 4.2. Why Ontologies but Not KGs?

While many people argue for KG superiority over ontology, ontologies are superior to KGs in *interoperability* and many *quality* measurements [20]. In fact, whatever approach is used to build a knowledge graph, the result will never be perfect [8]. Sources of imperfections are mainly because of *incompleteness, incorrectness and inconsistency* [45]. For example, If KGs are constructed from RSS feed or social media websites, there is a high probability that the knowledge will be noisy, missing important pieces of information or contains false information such as rumors. In addition, the accuracy of generated knowledge graph depends on the accuracy of the KG generation system.

Since the quality of generated KG is strongly dependent on the data quality of the knowledge source and the accuracy of KG generator, mapping the generated KG to reference ontologies will ensure quality and reliable representation of knowledge. Another reason for using ontologies over KGs is that most of the information systems in organizations are designed using domain-specific ontologies. Those systems cannot store and retrieve from KGs because *interoperability* issues would emerge if KGs are used.

## 5. Proposed Automatic ontology generation Framework

The proposed framework is inspired by the ontology generation life cycle developed in [2]. The framework consists of three main phases of the *Generation phase,* the *Refinement phase, and the Mapping phase* as shown in Figure 2. Each phase is discussed below in details.

### 5.1. Generation Phase

In the generation phase, the input to the framework is unstructured text corpus and the output is the preliminarily generated KG. The processes conducted in the generation phase are:

### 5.1.1. Data Cleaning

Data cleaning is a very important process. Without cleaning, irrelevant concepts and relations could deviate the reliability of results. Unstructured corpus might contain HTML tags, comments, social websites plugins, and ads. etc. Therefore, cleaning irrelevant information is necessary. Otherwise, we might find the term "Facebook" as one of the main concepts because it is repeated in many webpages.

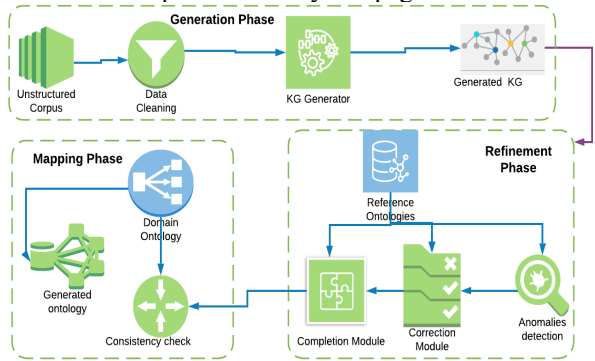

**Figure 2:** Proposed Automatic Ontology Framework

### 5.1.2. Knowledge Graph Generator

A KG generator can be used to generate the preliminarily KGs in the form of triples in RDF format (*subject, predicate, object*) or *(Resource, a Property, and a Property value)*. This generated graph includes the entities and relations with the corresponding confidence score. For example, the relation that USA is located near Mexico has a confidence score of 0.91, as shown below. This confidence score is not only retrieved from the corpus but also reinforced by previous knowledge stored in references ontologies. The generated KG can be also visualized on demand.

```
"results": [ {"id":
"eea16dfd5fe6139a25324e7481a32f89",
"result_metadata":{"confidence": 0.917}
```

## 5.2. Refinement Phase

As mentioned earlier, the most problematic issue with KG generation systems is that they cannot distinguish between reliable and unreliable knowledge source. Facts extracted from the Web may be unreliable, and the generated KG will be based on the information given in the knowledge source. So, deviation could emerge from the incorrect and incomplete ontological coverage of the generated KG [50]. The best approach to address these problems is to compare and complete the generated KG using prior reference knowledge.

### 5.2.1. Reference Ontologies

Reference ontologies are used to evaluate and verify the generated KGs. However, there is no single ontology that is considered the best reference for all domains. Accordingly, reference ontologies are selected based on the nature of problem of the generated KG in addition to the nature of domain ontologies themselves. For example, if the generated KG contains many general topics, *DBpedia* could be a perfect fit because *DBpedia* is the ontological form of Wikipedia [51]. In addition to reference ontologies, benchmark ontologies are needed to train the KG completion algorithms before they can be used in KG refinement.

### 5.2.1. Anomalies Exclusion

This step aims to exclude irrelevant, illogical, and unrelated nodes (concepts) and relations. There might be some nodes that are not connected to the rest of the graph. For example, if a political article contains some idioms such as "kick the bucket". Considering the bucket as a concept or entity is totally out of context. Most KG generators are creating concepts and relation along with confidence scores which represent the generator certainty about a concept or a relation. Therefore, the concepts or relations with low confidence scores should be removed.

### 5.2.3. Correctness Module

Based on literature review, insufficient research has addressed KG correctness. The primary source of errors in KGs are errors in the data sources used for creating KGs. *Association Rule Mining* has been used extensively for error checking and removing inconsistent axioms [52]. In the framework, the system learns about disjointness axioms or class disjointness assertion, and then apply those disjointness axioms to identify potentially wrong type assertions. For example, a school could be named "Kennedy", but a school cannot be a person (disjointness). So, a rich ontology is required to define the possible restrictions that cannot coexist [42]. DOLCE is a top level ontology that is rich with disjointness axioms [53].

### 5.2.4. Completion Module

This is the most important module in the framework as most of generated KGs are incomplete. KGs completion is called knowledge graph embedding which can be summarized as follows:
For each triple $(h, r, t)$, the embedding model defines a score function $f(h, r, t)$. The goal is to choose the $f$ which makes the score of a correct triple $(h, r, t)$ *is* higher than the score an incorrect triple $(h', r', t')$ [54].

Completion approaches could be classified into two categories: *Translational Distance Models* and *Semantic Matching Models.*

*Translational Distance Models,* such as the popular *TransE,* have been used extensively in many scientific research. However, *TransE* has defects in dealing with *1-to-N and N-to-N* relations [55]. While *Semantic Matching Models,* such as *RESCAL* and its extensions, link each entity with a vector to capture its latent semantics [56].

### 5.3. Mapping Phase

#### 5.3.1. Domain ontologies

Most information systems are based on domain ontologies. For example, in a hospital, there are fundamental healthcare ontologies. So, the generated KG must be mapped to fit in the domain ontologies to ensure the *consistency* and *Interoperability* generated KG with ontologies used in an organization.

#### 5.3.2. Consistency check

This step aims to solve interoperability issue in the KGs by checking whether all concepts and relations are consistent with the range and object-property in the target domain. For example, if $(x, p, y)$ indicates John ($x, the\ subject$) with property $(p, has\_lastName)$ of Robert ($y, the\ object$), then $has\_lastName$ makes John belongs to the domain class of ($Person, c$) or $(x, rdf:type, c)$. In the framework, we will extend the method developed by Péron et al. [57]. According to Péron, the domain inconsistency is defined as the occurrence of an object-property $p$ that does not belong to the containing domain. Similarly, the range inconsistency is the occurrence of object-property $p$ that does not belong to the definition range of $p$.

#### 5.3.3. The Generated ontology

After validating the consistency from the previous method, the KG is trimmed and transformed to a domain ontology. In this step, Super-subtypes classes are resolved and any relation that is inconsistent with the domain ontologies will be removed to ensure that the generated ontology could be easily integrated into the knowledge bases of the target organization.

## 6. Implementation

In this section, we will discuss the implementation details of the proposed framework in terms of the used algorithms, ontologies, and refinement methods.

**Data Cleaning:** For data cleaning, Python code was developed to parse webpages and search for HTML tags, irrelevant meta-tags and social media plugins, then, discard them from the input corpus. Data cleaning procedures are simplified in Algorithm 1.

| Algorithm 1: Cleaning Unstructured Corpus |
|---|
| **Input**: set of text file $T\{t_1, \ldots, t_i\}$ <br> **Output**: cleaned set of text files $T_C\{t_{C1}, \ldots, t_{Ci}\}$ <br> For each $t$ in $T$ <br>     If t extension is HTML <br>         Search for "\<p\>" tag or "\<span\>" <br>         Apply NLP to check sentence <br>         If tag forms a sentence <br>           If container tag contains no ads <br>             Add results tag to $T_C$ <br>         Break; <br>         Else <br>           Discard; <br>     Else if $t$ extension is RSS <br>         Search for "\<Description\>" or "\<Title\>" tags <br>         Add results tags to $T_C$ <br>     Else if $t$ extension is XML <br>         For each tag $tg$ in $t$ <br>           Apply NLP to the sentence <br>           If $tg$ contains a sentence <br>             Add $tg$ to $T_C$ <br>           Else <br>             Discard $tg$ <br>           End if <br>         End if <br>     Else <br>         If $t$ contains sentences <br>           Add $t$ to $T_C$ <br> End foreach |

**Selected Reference Ontologies**
To select the appropriate reference ontologies, we followed the criteria developed by [51] to find the most suitable knowledge graph for a given setting. Since we are adopting an organizational perspective, *YAGO* and *DOLCE* were the most suitable reference ontologies for the following reasons.
- *YAGO* currently has around 10 million entities and contains more than 120 million facts about these entities [58]. *YAGO* provides source information per statement. It also links classes to the WordNet Knowledge base and DBpedia.
- *DOLCE* is used for KG correction [53] which provides high-level disjointness axioms.

For our empirical analysis, we used WN18 and FB15k datasets, the most popular benchmark datasets built on WordNet [59, 60] for training and testing KG completion. These datasets serve as realistic KB completion datasets and are used for training completion module network.

**KG Generator:** to ensure best results, we used third-party KG generators to avoid the time and resources waste in building generic KGs (we recognize that this approach may not provide decent accuracy similar to those third-party solutions). *Neo4j* or *Watson* are third-party services that achieve outstanding performance in generating KGs from text. They offer KG generation as a service using APIs or special browsers.

**Anomaly Exclusion**: The first step is to exclude any concept or relation with a confidence score less than 0.3. The low confidence score means that the KG generator could not find enough evidence from the corpus nor from reference ontologies to support the concept or relation. Implausible links, such as RDF:sameAs assertion between a person and a book, can be identified based only on the overall distribution of all links, where such a combination is infrequent [61]. For concepts and relations with confidence score between 0.3 and 0.5, *Local Outlier Factor*[2] is applied to check their validity.

**KG Error Correction:** We used first-order logic to check the erroneous relations and associations by checking if each class has any relations with other classes found in the disjointness axioms associated with each class. For example, assume that the KG misrepresented "Columbus" city as a person but it has a *located_in* property. However, the class *Person* disjoints with *located_in* which belongs to the class *Location*. *DOLCE* ontology contains hundreds of axioms, that combines many (non-trivial) formalized ontological theories into one theory [52]. Aside from axioms, each property that is associated with each instance is validated using *YAGO* to check if the data is outdated or misrepresented. For example, imagine KG represented "Einstein" correctly as a scientist but incorrectly related him with Biology. In this case, *YAGO* would be used as reference for checking error beyond class types. The error correction process is summarized in Algorithm 2 along with SPARQL queries. The *DOLCE* axioms $A$ is generated using the following SPARQL query.

```
A= {PREFIX DOLCE
SELECT DISTINCT ?subject , DISTINCT ?
object ? property WHERE{ {?subject
DOLCE: type[]} UNION {[] DOLCE: type[]}
}}
```

| Algorithm 2: KG error correction |
|---|
| **Input**: $G\{g_1, \ldots, g_i\}$ is the generated KG consisting of triples and Let $A\{a_1, \ldots, a_i\}$ be a set of DOLCE axioms<br>**Output**: corrected KG $G_C \{G_{C1}, \ldots, G_{Ci}\}$<br>// search for disjoint axioms for each class<br>Foreach g in $G$.subject.Type<br>  Foreach a in A<br>    If (g rdf:disjoinWith(a))<br>      //search if class g has a relation with the in the graph G<br>      `PREFIX  G`<br>      `SELECT ?property ?value As e`<br>      `WHERE{  ?g rdf:SubclassOf a .`<br>      `?d rdf:onProperty a.}`<br>      //if erroneous relation found, delete from G<br>      If (e is not null)<br>        `DELETE {?property?value}`<br>        `WHERE { ? property`<br>        `rdf:type e. ? property`<br>        `rdf:type a}`<br>        Update $G$<br>      End if<br>    Else // no disjoint detected<br>      // use YAGO to check for erroneous info.<br>      `PREFIX  YAGO`<br>      `SELECT ? property ?value ?`<br>      `subject ? object  WHERE{ ?`<br>      `rdf:predicate g }`<br>      If (property | value| subject| object!=g.resources)<br>        Update $G$;<br>    End if<br>  End if<br>End Foreach<br>End Foreach |

**KG completion:** there are many models that have been used for KG completion. Among those models, Complex Embeddings (*ComplEx*) [62] has achieved promising results in comparison to other methods [63, 64]. *ComplEx* is considered the simplification of *DistMult* [65]. *ComplEx* uses tensor factorization to model asymmetric relations. *ComplEx* is implemented as a part of an embedding methods project on GitHub used for graph completion tasks[3]. The project contains *ComplEx* as one of six powerful graph embedding methods such as *TransE* and *HolE*. We are using the same settings created by the project in terms of number of epochs, batch sizes, and optimization function.

*ComplEx* is based on Hermitian dot product, the complex counterpart of the standard dot product between real vectors. In *ComplEx*, the embedding is complex which means that it has a real and imaginary value. The entity and relation embeddings $(h, r, t)$ no longer lie in a real space but a complex space, say $\mathbb{C}^d$. The score of a fact $(h, r, t)$ is defined as

$$fr(h,t) = Re(h^\top diag(r)\bar{t}) = Re\left(\sum_{i=0}^{d-1}[r]_i \cdot [h]_i[\bar{t}]_i\right)$$

where $\bar{t}$ is the conjugate of $t$ and $Re(.)$ which means taking the real part of the complex value. So, $fr(h,t)$ produces asymmetric relations that represent the acceptance of different scores depending on the order of entities involved [62].

---

[2] The function on GitHub, LOF (https://github.com/ronak-07/Outlier-Detection-LOF)

[3] The project on GitHub, Graph Embedding project (https://github.com/mana-ysh/knowledge-graph-embeddings)

## 6.1. Domain ontologies consistency

In order to process the generated KG, it must be consistent with the target domain ontologies. Otherwise, we cannot store and retrieve knowledge from the domain Knowledge base. So, given that $On\{On_1,..,On_j\}$ are the domain ontologies of the organization $O$, we will treat the generated KG as a temporarily ontology named $OKG$. For each concept $c$ in the ontology $OKG$, the number of domain inconsistencies $\varepsilon(c)$ is calculated as the sum of the differences between the properties in the target ontology $On$ and the properties in the generated ontology $OKG$ which is defined by the following relation:

$$\varepsilon(c) = \sum_{p=1}^{j} |On(p)| - |OKG(p)|$$

Consequently, if $OKG$ has $m$ number of concepts, the overall inconsistency $\varepsilon$ is the sum of domain inconsistencies for the $m$ concepts exist in the $OKG$. Consequently, the final resulting ontology ($domainOnT$) is the generated $OKG$ excluding the total domain inconsistences $\varepsilon$. $domainOnT$ is represented in the following relation

$$\varepsilon = \sum_{c=1}^{m} \varepsilon(c), \quad domainOnT = OKG - \varepsilon$$

## 7. Illustrative Example

Company $D$ wants to store the latest financial updates of the Fortune 500 companies in its knowledge base. The company is using *WordNet* knowledge base and *DBpedia* domain ontologies. The company tried our framework to store the latest ranks, performance, and revenue of the Fortune 500 into the company *WordNet*.

***KG Generation:*** The company uploaded an Excel sheet and a detailed webpage about the performance of each Fortune 500 company[4,5]. After uploading the documents, the webpage header, footers, and unrelated content were removed using Algorithm1. Then, text was converted to a KG using *Neo4j* KG generator. The resulting KG contained information about the latest Fortune 500 companies and is shown in Figure 3. We limited the resulting nodes in the figure to 300 nodes to be readable, as the original graph contained more than 3000 nodes. Each concept in the figure was colored using a unique color. For example, the company name was colored light blue, business focus was colored orange, revenue change in dark blue, and rank in beige.

***KG Refinement:*** The initially generated KG was certainly not perfect. At first, anomalies were extracted by removing scattered unrelated nodes on the top left of the graph because their confidence scores were very low, and they were not connected to the rest of the graph. In the next step, the KG was corrected for erroneous relations and properties. As shown in the graph, Facebook was mistakenly represented as a motor company instead of a tech. company. Delta Air Lines was also misrepresented as a retail company. By applying Algorithm 2, no disjointness constraints were found. Both companies had the correct properties that belong to the class *Company*. However, after using *YAGO* as a reference ontology, the error in the two companies were easily detected and corrected.

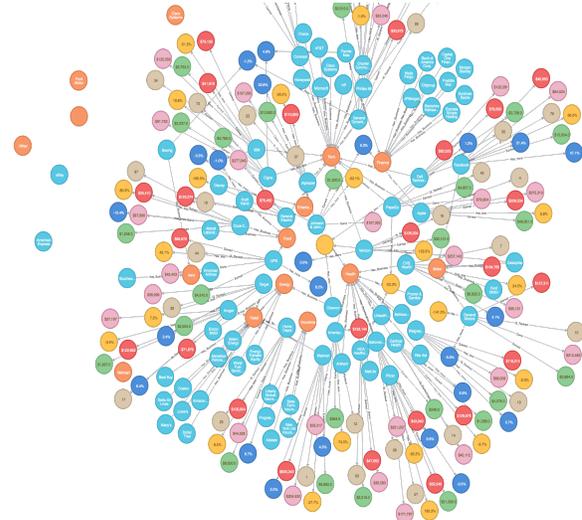

**Figure 3:** Initially generated Knowledge graph of the Fortune 500 companies.

Unfortunately, the information provided in the uploaded files was incomplete, and only 70 Fortune companies were assigned to business focuses (for example, Apple was assigned to "Technology" while others had empty business focus). So, the remaining unassigned companies must also be assigned to a business focus. To complete the missing relations and entities, *ComplEx* was applied to the KG. However, to avoid conflict in naming business focus, *ComplEx* was used to assign the whole KG, even for those companies who were already assigned. The 70 companies were used to check if the new business focus was close or the same as old business focus using *Azure Search Service REST API* [6].

---

[4] https://fxssi.com/top-10-profitable-companies-world
[5] https://www.someka.net/excel-template/fortune-global

[6] Azure Search Service REST API :docs.microsoft.com/en-us/rest/api/searchservice/create-synonym-map

*Consistency with Domain Ontologies*

While the generated KG is now refined and completed, it is not necessarily consistent with domain ontologies. Generated concepts and relations are based on the initially uploaded documents. In our example, the generated KG has the following concepts: "*Rank, Company Name, Employees, Previous Rank, Revenues, Revenue Change, Profits, Profit Change, Assets, Business, and Market Value.*" However, Company D used DBpedia as a domain ontology that does not have *Previous Rank, Revenue Change, Profit Change* as properties, so these properties and related attributes were removed. Therefore, the final generated ontology is the intersection between the concepts and properties in the generated KG and DBPedia.

## 8. Conclusion

Many information systems depend on ontologies as reliable representation of knowledge. However, generating ontologies is a tedious and costly process that require domain experts' intervention. Ontologies are hard to evolve and might be outdated. Automatic ontology generation would help ontologies evolve and save on the cost and time of ontology creation and maintenance. However, automatic ontology generation from heterogeneous text sources is still an open area of research. This research aims to develop a domain independent automatic ontology generation framework from unstructured text corpus using KGs. The framework will allow organizations to convert new knowledge to consistent ontological form. The framework maps KG into target domain ontologies using *ComplEx* for KG completion and reference ontologies for KG refinement. Future research will include assessing the validity of the system across different domains based on the *Tasks Technology fit* theory [66] as well as evaluating the framework using different semantic accuracy measures.